# An Efficient Deep Convolutional Neural Network Model For Yoga Pose Recognition Using Single Images

Santosh Kumar Yadav[a,b,∗], Apurv Shukla[c,∗], Kamlesh Tiwari[d], Hari Mohan Pandey[e], Shaik Ali Akbar[b,a]

[a]*Academy of Scientific and Innovative Research (AcSIR), Ghaziabad, UP-201002, India*
[b]*Cyber Physical System, CSIR-Central Electronics Engineering Research Institute (CEERI), Pilani-333031, India*
[c]*Department of EEE, Birla Institute of Technology and Science Pilani, Pilani Campus, Rajasthan-333031, India*
[d]*Department of CSIS, Birla Institute of Technology and Science Pilani, Pilani Campus, Rajasthan-333031, India*
[e]*Department of Computing and Informatics, Bournemouth University,United Kingdom*

## Abstract

Pose recognition deals with designing algorithms to locate human body joints in a 2D/3D space and run inference on the estimated joint locations for predicting the poses. Yoga poses consist of some very complex postures. It imposes various challenges on the computer vision algorithms like occlusion, inter-class similarity, intra-class variability, viewpoint complexity, *etc*. This paper presents YPose, an efficient deep convolutional neural network (CNN) model to recognize yoga asanas from RGB images. The proposed model consists of four steps as follows: (a) first, the region of interest (ROI) is segmented using segmentation based approaches to extract the ROI from the original images; (b) second, these refined images are passed to a CNN architecture based on the backbone of EfficientNets for feature extraction; (c) third, dense refinement blocks, adapted from the architecture of densely connected networks are added to learn more diversified features; and (d) fourth, global average pooling and fully connected layers are applied for the classification of the multi-level hierarchy of the yoga poses.The proposed model has been tested on the Yoga-82 dataset. It is a publicly available benchmark dataset for yoga pose recognition. Experimental results show that the proposed model achieves the state-of-the-art on this dataset. The proposed model obtained an accuracy of 93.28%, which is an improvement over the earlier state-of-the-art (79.35%) with a margin of approximately 13.9%. The code will be made publicly available.

∗Authors contributed equally.

*Email addresses:* santosh.yadav@pilani.bits-pilani.ac.in (Santosh Kumar Yadav), f20180405@pilani.bits-pilani.ac.in (Apurv Shukla), kamlesh.tiwari@pilani.bits-pilani.ac.in (Kamlesh Tiwari), profharimohanpandey@gmail.com (Hari Mohan Pandey), saakbar@ceeri.res.in (Shaik Ali Akbar)

## 1. Introduction

The pose of a person represents a particular orientation which is essentially related to understanding various physical and behavioral aspects. Human pose recognition has numerous applications ranging from human-computer interaction, animation, gestural control, virtual reality, sports, sign language recognition, *etc*. [1]. It is challenging due to factors such as a large variety of human poses, rotation, occlusion of limbs, body orientation, and large degrees of freedom in the human body mechanics [2, 3].

With the growth of online media, the amount of video and image databases are tremendously increasing on web platforms like YouTube, Netflix, Bing, *etc*. The computer vision community has made use of this and many works exploited the indefinite source of data available on the web to build large pose recognition datasets comprising a variety of postures, thus enhancing the application scope of recognition algorithms. Despite several efforts of building large-scale datasets, not many datasets deal with complex human poses especially that involved in performing yoga asanas. For example, Andriluka *et al*. [4] proposed MPII dataset containing approximately 25,000 images with over 40,000 people. Each image is extracted from a YouTube video. The dataset covers 410 specific categories of human activity and 20 general categories. Though the aforementioned large-scale dataset has introduced pose diversity, in terms of the complexity of human pose, they are nowhere close to the complex body postures of yoga exercises [5].

Yoga is popular across the world as a safe and effective exercise [6]. It comprises body postures of various complexities. Yoga postures offer some of the complex body orientations that can be hard to capture from a single viewpoint. The complexity further increases with the occlusions, and changes in image resolutions. Figure 1 presents some of the challenging yoga postures consisting of inter-pose similarity, intra-pose variability, different styles of doing a particular asana, self-occlusion, and synthetic images from the Yoga-82 dataset [5]. Because of these complexities, generating fine annotations of body keypoints is not feasible and therefore, the existing state-of-the-art keypoint detection-based approaches may not be suitable for yoga pose recognition [5].Due to the inherent challenges in the pose, current pose estimation methods are not able to correctly predict the pose on the yoga asanas [1].

Few works, like [7, 8], proposed video-based yoga recognition systems. The dataset of [7] consisted of videos for 6 asanas (*i.e.* bhujangasana, padmasana, shavasana, tadasana, trikonasana, and vrikshasana) recorded with an RGB webcam, whereas [8] used Kinect to capture depth maps for the recognition of 12 yoga poses. However, the dataset they built contains relatively simple yoga poses with less occlusion, less number of classes, and does not offer many challenges to the learning algorithms. However, these works, [7, 8] lack in terms of generalization ability. Recently, Verma *et al*. [5] introduced Yoga-82 dataset of static images with 82 yoga pose classes. It consists total of 28.4k images. The dataset has three levels of the hierarchy. The Top, mid, and class level hierarchies consist of 6, 20, and 82 classes, respectively. The images were collected from various online sources and are of different resolutions, illuminations, viewpoints, and occlusions. Moroever, in the dataset there are a few synthetic yoga images as well, dealing with silhouette and cartoon based poses. We utilize the Yoga-82 dataset as a testbed for our proposed network.

In the last decade, convolutional neural networks (CNNs) have achieved significant progress on pose estimation, object detection, and semantic segmentation tasks.However, existing pose estimation approaches, like OpenPose [9], HRNet [10], PifPaf [11], Fast Pose [12], *etc*., usually have a low performance rate and less robustness when applied to complex yoga postures. Due to different challenges in yoga poses, the pose estimation based approaches rather fail to predict the self-occluded body joints.The vision community has made significant progress in semantic segmentation and object detection [13]. For example, Mask R-CNN [13] extends the Faster R-CNN [14] by adding a branch for predicting an object mask in parallel to the existing branch of bounding box recognition. It has been tested for three tasks *i.e.* instance segmentation, person keypoints detection, and bounding box detection for the object. Motivated by these recent advances, we utilized the object detection and semantic segmentation-based approach in our proposed approach for refining the yoga poses.

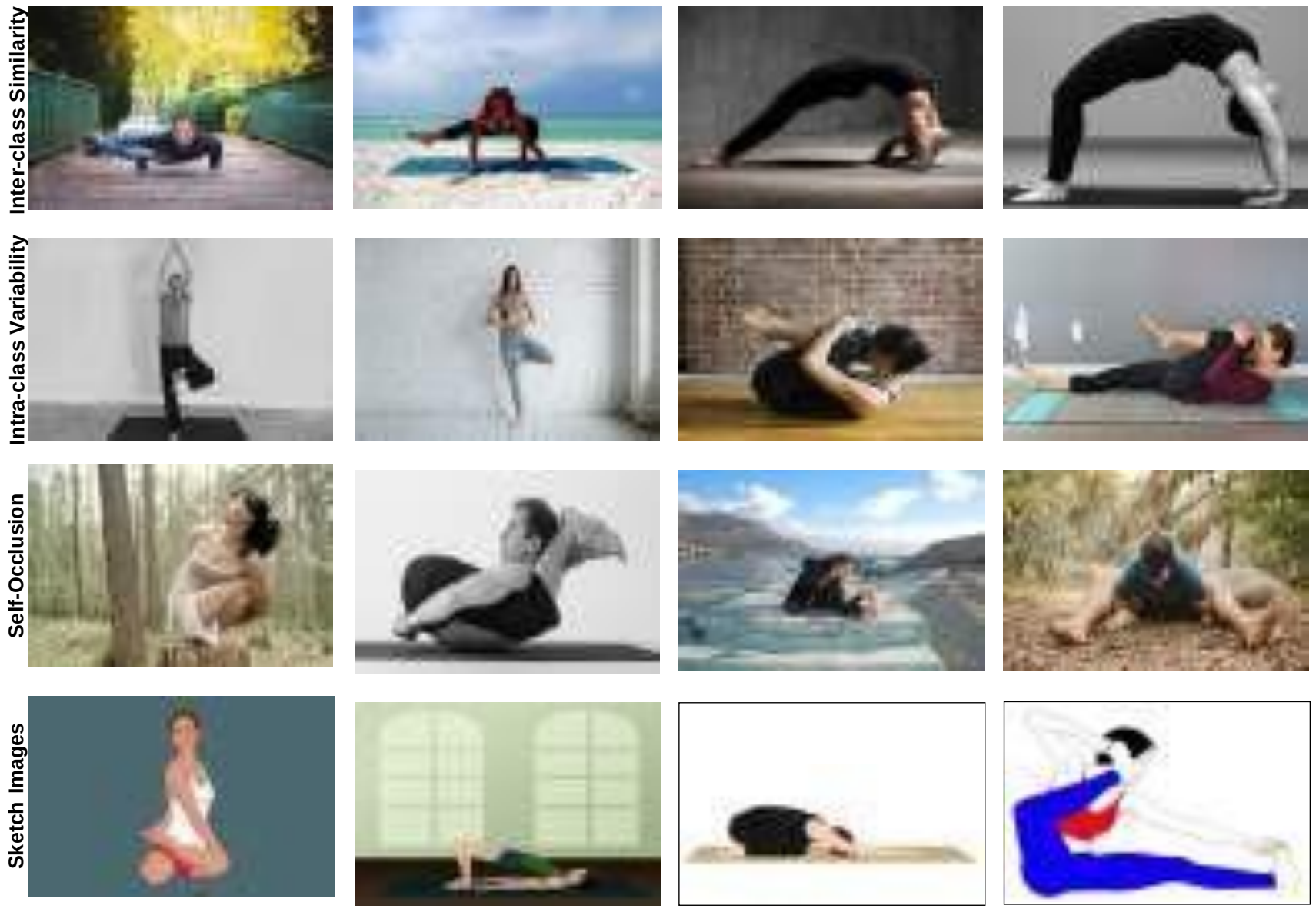

Figure 1: Complexity in Yoga poses due to factors like inter-pose similarity, intra-pose variability, self-occlusion and cartoon images in the Yoga-82 [5] dataset.

In this paper, a yoga pose recognition network is proposed to efficiently recognize complex postures using static images of different yoga asanas. The proposed approach aims at directly predicting the pose of yoga being performed by a practitioner. First, the pose image is refined using a segmentation framework, which detects the person in a given image and segments it along with bounding box annotations. For the object detection and segmentation, we followed a approach inspired from Mask R-CNN [13]. The predicted segmentation masks on the region of interest (ROI) are used for extracting bounding boxes. The extraction of ROI from 2D images helps in reducing the irrelevant background information from input images which helps in better learning of proposed YPose network. The refined images are then inputted to a deep neural network, which consists of the EfficientNets [15] backbone. The EfficientNet architecture is showed to outperform the state-of-the-art CNN networks including DenseNet [16]. The EfficientNet maintains an optimum scaling which achieves improved performance with balanced computation costs. The proposed model utilizes the B4 variant of EfficientNets.We further modify the network architecture by adding a number of dense refinement blocks adapted from the work of DenseNets, followed by global average pooling and fully connected layers in an end-to-end manner. From our study, it is evident that by the addition of dense refinement blocks there is an improvement over the baseline variants.The proposed YPose network has been evaluated on a publicly available benchmark dataset named Yoga-82 [5]. Figure 2

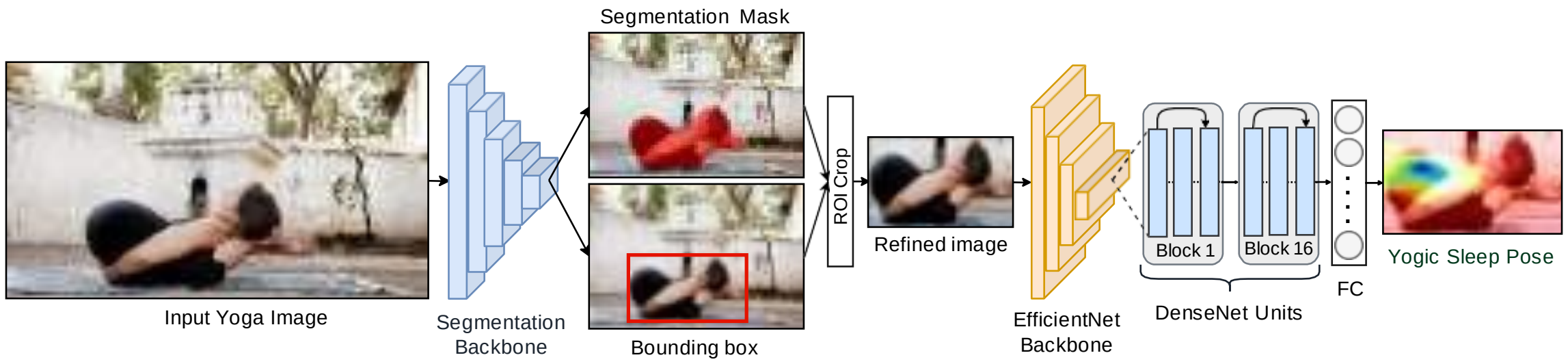

Figure 2: Naive representation of the approach for building the proposed network. As shown in the figure, the input image is first refined using a segmentation based network and passed to the proposed CNN architecture consisting of EfficientNet [15] backbone and dense refinement blocks for feature extraction. Finally, the yoga is classified using a fully connected layer. From the activation map of the final layer it can be observed that the proposed model is able to learn meaningful representation of complex and occluded yoga posture.

presents a naive representation of our proposed yoga pose recognition method.

In particular, the major contributions of the proposed model are as follows.

- We propose a novel approach for yoga pose recognition using single images. The proposed network is shown to extract high-level pose specific features to recognise the complex yoga poses in an end-to-end manner. Our approach aims to filter the irrelevant information of yoga images to learn informative representations of a yoga pose. The proposed network is robust to noise caused by challenging backgrounds, body keypoint occlusions and complex yoga poses.

- We combine image segmentation based approaches to first segment and locate the region of interest (ROI) in the input RGB image. The extracted segmentation masks and ROI are then used to refine the image to filter the background contexts that are not informative to the yoga pose recognition task. The refined images are passed to a deep CNN architecture, further modified to learn fine-grained pose features. Finally, we add global average pooling and fully connected layers to obtain accurate pose predictions.

- Due to the unavailability of a large scale yoga dataset, we exhaustively conducted computer simulation experiments on the Yoga-82 dataset to evaluate the robustness of our proposed approach.We applied the baseline architectures of EfficientNet, B0, B4, B5 and obtained accuracies of 86.76%, 89.25% and 89.46%, respectively on third level hierarchy. The proposed approach was not evaluated for the deeper variants of the aforementioned architecture to avoid the high computational cost due to the increased number of floating-point operations (FLOPS). Instead, to improve the model performance with current backbone parameters, we propose to add several dense refinement blocks adapted from the architecture of DenseNet [16] were used. The YPose network achieved an accuracy of 93.28%.

- The proposed YPose network has been compared with the state-of-the-art results on the Yoga-82 dataset. The previous state-of-the-art on this dataset was 79.35% and 93.47% for the Top-1 and Top-5 accuracies for the 82 classes.Our proposed model has achieved 93.28% and 98.04% for Top-1 and Top-5 accuracies, respectively, which outperforms the previous state-of-the-art with a margin of 13.9%.

- The proposed network is lightweight and robust to complex yoga poses and the associated challenges to them such as illumination, occlusion. From the results, we observe that the network learns pose specific features and is able to address the inherent challenges in a particular yoga pose such as inter-pose similarity and intra-pose variability. The total number of parameters of our proposed model is 22.68 M, which are also comparable with the model of Yoga-82 consisting of 22.59 M parameters. The proposed model can easily run on the freely available GPUs provided by the Google Colaboratory.

Section 2 presents the literature study of pose estimation and yoga pose recognition methods. In Section 3, the proposed methodology is described, followed by the process of pose extraction, model description, and training the network. The training outcomes and final results are discussed in Section 5 and it shows the potential of our technique for real-world applications. It contains the dataset details, experimental results, performance evaluation of the proposed network, and discussion and comparison. Finally, in Section 6, the concluding remarks and future research directions are presented.

## 2. Related Works

In this section, we review the recent pose estimation and yoga pose recognition literature.

### *2.1. Pose Estimation*

In the last decade, the performance of computer vision algorithms has seen a significant improvement in terms of complex human pose estimation. It has attracted the attention of many researchers. Conventional pose estimation methods focus on detecting joint keypoints, and consequently estimating poses, from the target images [17, 18, 19]. Cao *et al*. [9] presented, OpenPose, a real-time approach to detecting the 2D pose keypoints of multiple people in an image. The approach uses part affinity fields (PAFs), to learn to associate body parts with individuals in the image. Similarly, PifPaf [11] proposed a pose estimation method that used part intensity fields and part association fields to first localize the body parts and then associate these parts to the human body pose. However, they fail to give accurate keypoints predictions for the yoga poses involving inversion, self-occlusion and other complex styles. Ni *et al*. [20] proposed a human pose detection system by acquiring kinematic parameters of the human body using multi-node sensors. However, the multi-node sensors, generally, might not be available in common households. HRNet [10] utilized both low and high-resolution representations of the image encodings for forming more stages and performing multiscale fusion between sub-networks. UniPose [3] proposed a unified framework for the human pose estimation using contextual segmentation and joints localization. Fast-Pose [12] proposed a fast pose distillation model learning by training lightweight pose neural networks with a low computational cost. Recently proposed, BlazePose [1] made an assumption that the face or at least the head of the user must be visible for pose estimation. However, these assumptions are unrealistic and are violated under typical scenarios of yoga poses. Moreover, most of the algorithms have been evaluated on datasets that are relatively simple in terms of human pose complexity, activity diversity and thus, do not present enough challenges to the recognition algorithms, limiting their application scope.

### *2.2. Yoga Pose Recognition*

Few works have been proposed on yoga pose recognition. For example, Chen *et al*. [21, 22, 8] proposed a self-training system for yoga pose recognition and correction using Kinect depth sensors. In [21], the contour information, skeletal features, and body coordinates were extracted using two

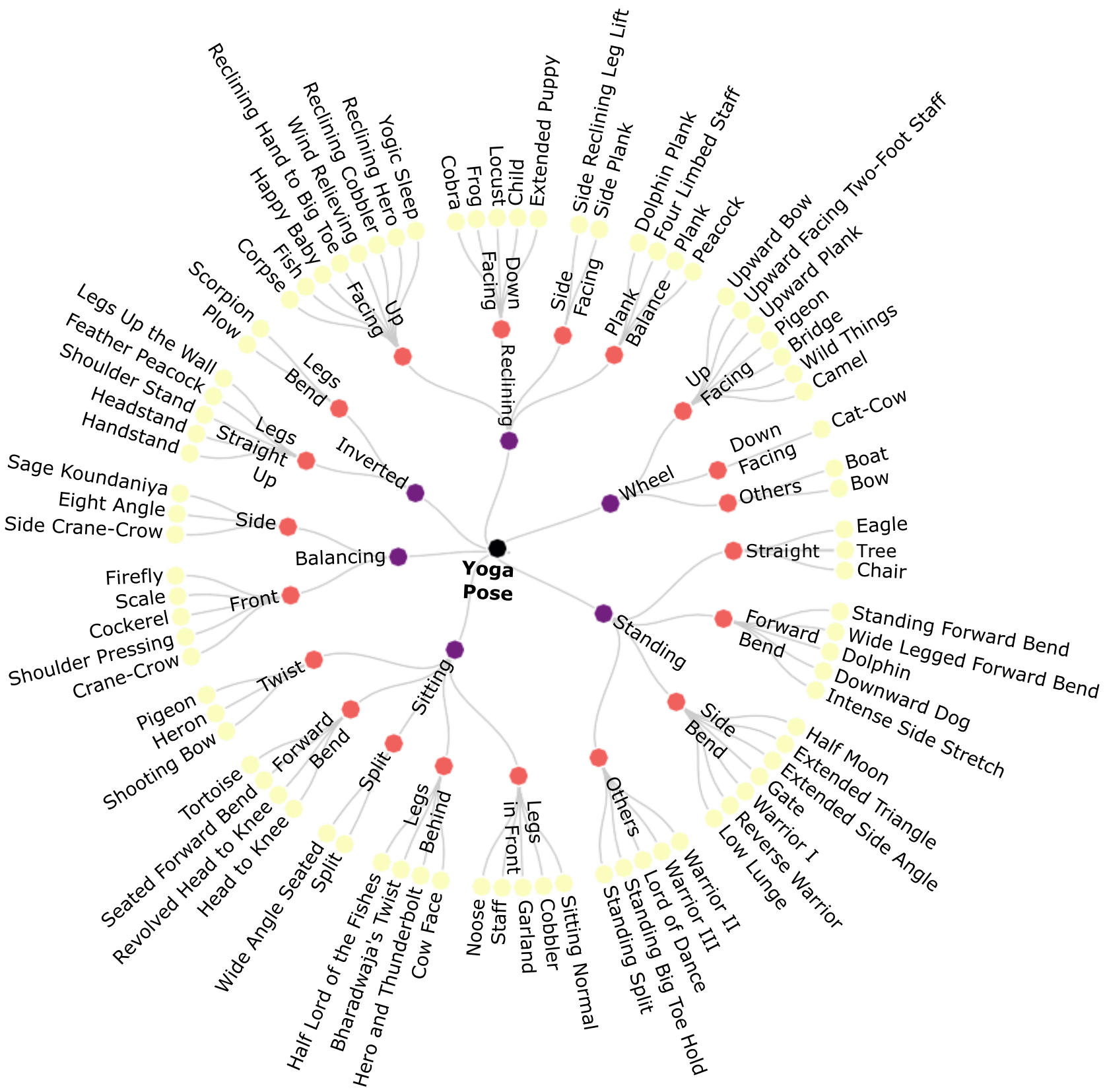

Figure 3: Three level hierarchy of the yoga poses in the Yoga-82 dataset [5]. Theplum,orange, andyellowcolor dots represents first, second and third level hierarchy consisting of 6, 20, and 82 classes, respectively.

Kinect depth sensors by placing them perpendicular to each other. In [22], the model first extracted the contour information and then applied the star skeleton method for pose representation. The model was tested for twelve different yoga postures, where, five yoga practitioners have performed each asana five times. [8] extracts feature using Microsoft Kinect and OpenNI library for recognizing twelve poses. However, they made separate models for each asana and calculated features manually. Moreover, depth-sensor cameras may not be available in common households. Maddala *et al*. [23] presented a 3D motion capture system for yoga poses in videos using joint angular distance maps (JADMs) and CNNs. They evaluated their model performance on a self-collected dataset using 9 cameras, 8 IR, and 1 RGB video camera for 42 yoga poses, and on two publicly available datasets, namely CMU [24] and HDM05 [25]. Yadav *et al*. [26] proposed a yoga recognition system to classify six asanas. They used OpenPose [27] to detect eighteen keypoints of the human body and passed them to a hybrid deep learning model consisting of CNN and LSTM. Likewise, Jain *et al*. [28] proposed a yoga pose recognition system using 3D-CNNs for ten yoga poses.

However, these works involve a yoga dataset with a less number of images or videos and do not consider the vast variety of poses. They lack generalization capabilities and are far from recognizing complex yoga poses. Recently, Verma *et al*. [5] introduced fine-grained hierarchical yoga pose recognition dataset consisting of 28.4k images for 82 classes of yoga *asanas*. The dataset consists of a three-level hierarchy as shown in Figure 3. They tested different variants of CNN architectures *i.e.* ResNet [29], DenseNet [16], MobileNet [30], and ResNext [31]. However, the model performance was quite low because the dataset consists of inherent challenges like inter-pose similarity, intra-pose variability, occlusion, and presence of the synthetic images involving silhouette and cartoon poses. This paper presents a novel approach to recognize complex yoga asanas in single images.

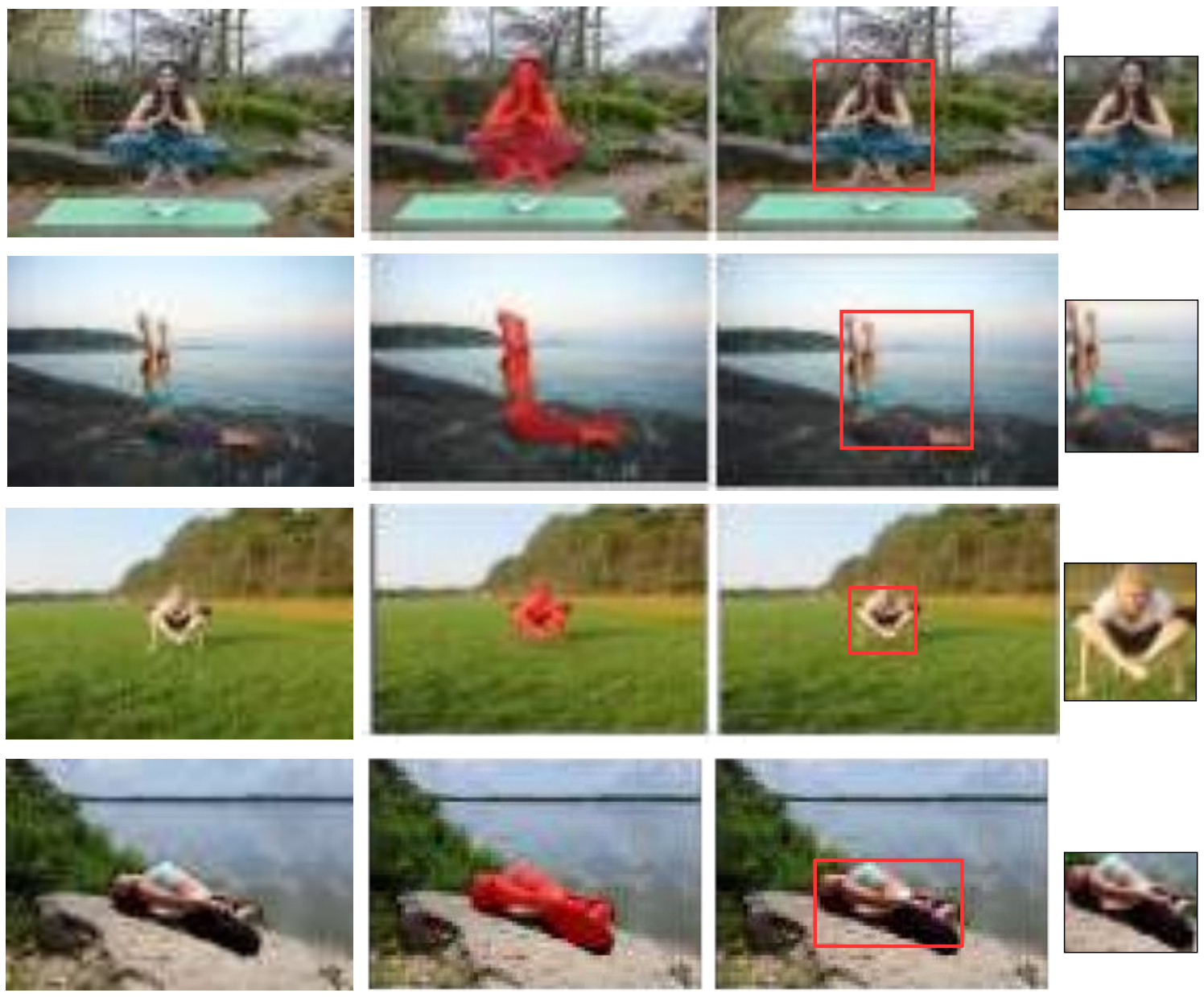

Figure 4: ROI segmentation of the yoga pose images. First column presents the original images from the Yoga-82 [5] dataset. The second column presents the mask generated on the original images. The third column shows bounding boxes generated for the person instance. Finally, the fourth column presents the obtained refined pose images.

## 3. Proposed Methodology

This section presents the proposed methodology of the proposed model. The proposed network recognizes complex yoga poses efficiently from RGB images. The proposed model consists of four main components *i.e.* ROI segmentation, EfficientNet [15] backbone, dense refinement blocks, and fully connected layers. In the first component, instance segmentation and object detection is applied to the original images to extract the ROI from the in-the-wild yoga images. We define ROI as the region of the image where the yoga practitioner is present. In the second component, the refined images are passed to the EfficientNet [15] based backbone to compute the spatial features. In the third component, dense refinement blocks are applied to obtain more diverse features. Finally, in the fourth component, global average pooling and fully connected layers are applied to get the prediction scores for the first, second, and third-level hierarchy of the yoga poses. Figure 5 illustrates the overall pipeline of the proposed model. This is further described in detail in the following subsections.

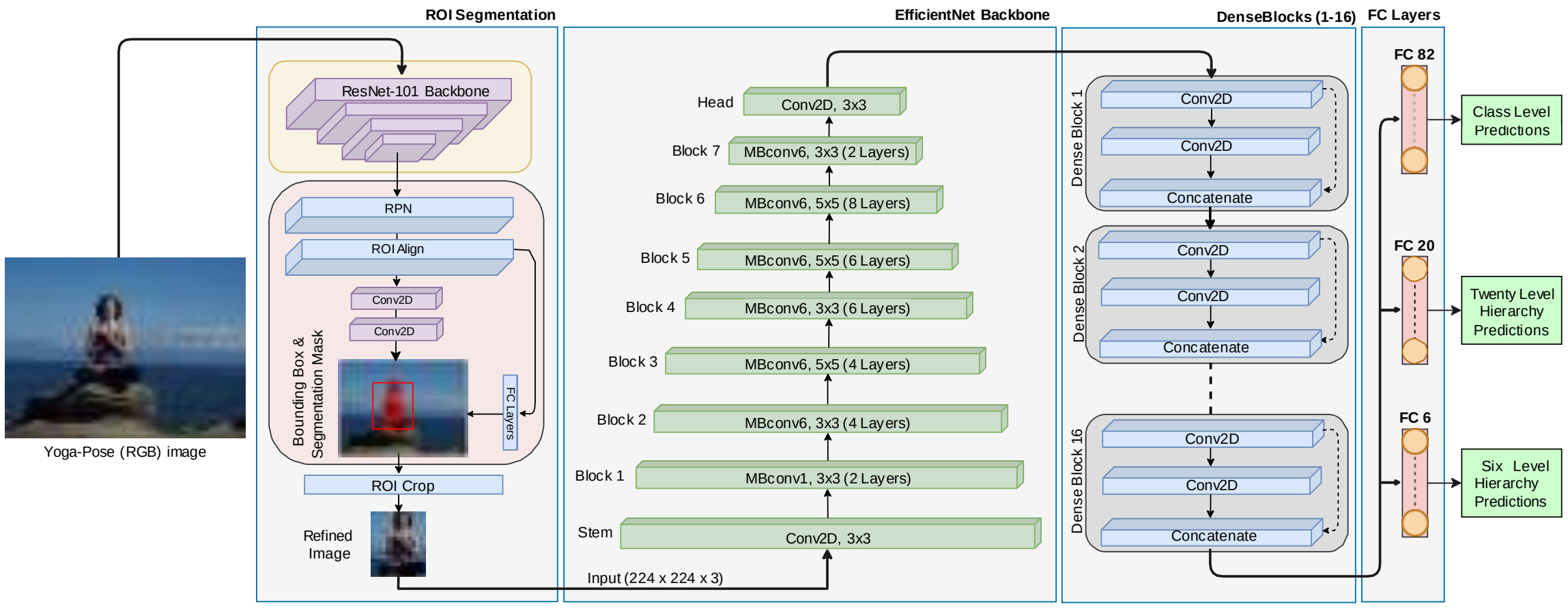

Figure 5: The schematic representation of the proposed model. The proposed network consists of four steps. In the first step, ROI segmentation is performed on the input image using segmentation network, RPN, and ROI align. Using this we obtained object mask, bounding box and object label. Next, refined images are inputted to the EfficientNet [15] backbone network followed by several refinement block units to calculate diverse features. Finally, fully connected layers give output for the three-class hierarchical predictions for yoga poses.

### 3.1. *ROI Segmentation*

This subsection describes the proposed approach of ROI segmentation and extraction. Apart from the various inherent complexities of the yoga poses due to self-occlusion, interclass similarities, intraclass variabilities, illumination Yoga-82 [5], dataset consists of synthetic images with cartoon sketches and silhouette poses and images where the person is performing yoga in the wild (*e.g.* hills, outdoor terrain) that makes the yoga pose recognition even more challenging. To deal with these challenges associated with the practitioner's environment, we propose an ROI segmentation approach for yoga poses as shown in Figure 4.

The ROI segmentation model is used to classify every pixel location to perform segmentation for each object instance in the image. It typically consists of ResNet-101 [29] backbone, region proposal network (RPN), and ROIAlign layers. The Resnet-101 backbone is used to extract the features maps from the input image. The RPN layer runs over the image to obtain bounding box coordinates for the region proposals *i.e.* regions in the feature map containing 'Person' as an instance. Further, the outputs from the RPN layer are fed into the ROI align layer to obtain segmentation masks for each object present in the image.

This generates bounding box coordinates along with high-quality segmentation masks for each object instance in the image. The ROI segments are cropped from the images. As the size of the generated annotation was different for each image, all these images are resized to 224 × 224 size. However, the synthetic images were passed directly to the CNN architecture as these were leading to erroneous ROI predictions.

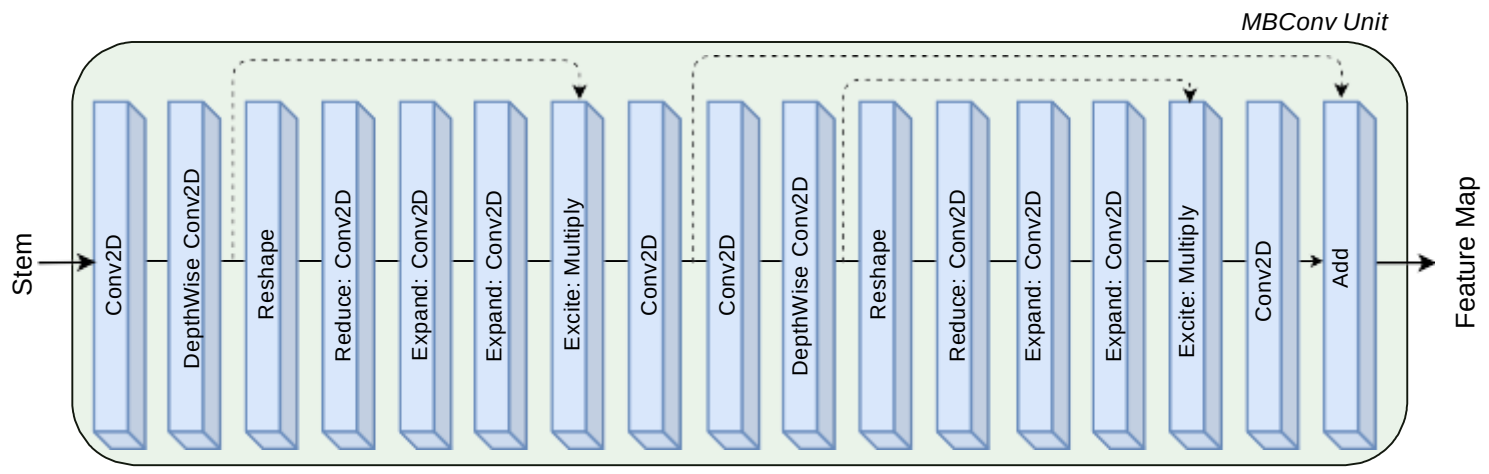

Figure 6: The representation of one MBconv unit, which are the basic building blocks of the EfficientNet [15] backbone.

### *3.2. Proposed Model*

The refined images of size 224 × 224 obtained from the ROI segmentation are passed to the deep CNN architecture. The deep CNN architecture consists of the EfficientNet [15] backbone and dense refinement block units, adapted from the architecture of DenseNet [16] followed by global average pooling and fully connected layers. The EfficientNet architecture comprises seven blocks with varying filter size, strides, and the number of channels. Each block consists of the mobile inverted bottleneck unit, namely MBConv [30]. The MBConv units use depthwise convolution and add the squeeze and excitation optimization. Figure 6 represents one such MBConv unit. In the expansion stage, we are increasing the network width by increasing the number of channels. Next, the depthwise convolution operations are performed and finally, in the squeeze and excitation stage, the global features are extracted.

Starting from the baseline variant B0 of EfficientNets, the architecture is systematically scaled according to the width and depth parameters to the deeper variant B7. As demonstrated in [15], for the models deeper than B4, the performance does not improve significantly with the increase in the number of network parameters.

Settings used for the depth parameter, *d* and width parameter, *w* were 1.8 and 1.4, respectively. The number of convolutional layers and filters were modified according to the scaling parameters ($d$ and $w$) as $f_{inc} = w \times f_{input}$, where $f_{inc}$ denotes modified number of filters and $f_{input}$ denotes number of input filters. The depth divisor ($d_{div}$) is set to 8 for scaling operations and the value of minimum depth, $d_{min}$ is kept equal to $d_{div}$. For scaling the width of the network we modify the filters as Equation 1, where $filter_{scaled}$ is the new number of filters.

$$filter_{scaled} = \max\left(d_{min}, \left\lfloor \frac{f + \frac{d_{div}}{2}}{d_{div}} \right\rfloor \times d_{div}\right) \quad (1)$$

For scaling the depth of the network we increase the number of MBConv units according to the depth coefficient, $d$ as $repeating\ units = \lceil d \times repeats \rceil$. Where repeats are the default number of repeating MBConv units which are further scaled according to the depth scaling parameter.

We conduct our initial experiments with the vanilla baseline B0, B4 and B5 of EfficientNets. From Table 2, we observed improvement in perfromance from B0 to B5. However, the Top-1 accuracy on the B5 baseline was not significantly increased on the addition of approximately 6 million parameters compared to baseline B4. Therefore, we preferred B4 instead of higher variants for the choice of model backbone.

The architecture of B4 serves as the backbone of the proposed model for feature extraction. With the current number of parameters, we further modify the network by adding several dense refinement block units to further improve the performance. These blocks are adapted from the architecture of DenseNet [32]. Each refinement block typically consists of several 2D convolution layers with varying number of filters. The number of these blocks to be added is a crucial hyper-parameter for the proposed approach. Figure 9 depicts a detailed analysis of this hyper-parameter. From the experiments we infer that 16 such units are used to achieve optimum and efficient performance.Figure 5 represents the detailed description of the modified architecture. There are considerable improvements in predictions after addition of the refinement blocks. Each block concatenates the features with the feature maps of preceding layers as shown in Equation 2.

$$x_l = F_l[x_{l-1}, ..., x_1, x_0] \tag{2}$$

where, $x_l$ denotes the feature vector of *lth* layer and $F_l$ denotes the corresponding mapping.

Throughout the proposed model, the swish activation function was used. We preferred the swish activation function over ReLU because of its superior performance for the complex image recognition tasks [33]. The smoothness of the swish activation function helps the deep neural network to optimize and generalize better. The learnable parameter $\beta$ was set to 1.0. If $f(x)$ denotes the swish activation funtion and $\sigma(x) = \frac{1}{1+e^{-x}}$ then,

$$f(x) = x \times \sigma(\beta \times x) \tag{3}$$

The output of the final block is passed as an input to the fully connected layers for the classification. The net output of the fully connected layers, $z^{(i)}$ is defined as $z^{(i)} = w_m \times x_m + w_{m-1} \times x_{m-1} + ... + w_1 \times x_1 + w_0 \times x_0$, where $w_j$ and $x_j$ are the weight and the learned feature vectors, respectively.

$$\hat{P}(class = i|c^{(j)}) = \Phi(c^{(j)}) \tag{4}$$

where, $\Phi(c^{(j)})$ denotes the Softmax function, as given in Equation 5.

$$\Phi(c^{(j)}) = \frac{e^{c(j)}}{\sum_{i=0}^{n} e_n^{c(i)}} \tag{5}$$

$\hat{P}$ denotes the probability of $j^{th}$ training example belonging to class $i$ predicted by the model for each six-level, twenty-level, and class-level hierarchies. The class that has the highest probability score is predicted by the network. The total number of parameter details of the proposed model is presented in Table 5.

### *3.3. Model Training*

The proposed model has been trained on the freely available GPU provided by the Google Colaboratory, ensuring that the network can be run using limited resources. All the experiments were developed using Keras and Tensorflow 2.0 backend. The train, validation, and test split consist of approximately 21K, 3.7K, and 3.7K images, respectively. The images were passed in mini-batches of size 32. Hyperparameter tuning was performed for the number of refinement blocks to be used in order to obtain optimal performance with balanced FLOPs. Categorical cross-entropy loss function was used to compile the network. For training the architecture, ADAM optimizer was used with a learning rate of 1.0e-5. Default settings were used for the hyper-parameters $\beta_1$ and $\beta_2$. The model was trained for 50 epochs. Each epoch took approximately 13 minutes to complete.

The network has been run on the different levels of the hierarchy of the dataset. Figure 7 illustrates the accuracy and loss variation for the training and validation with the number of epochs.

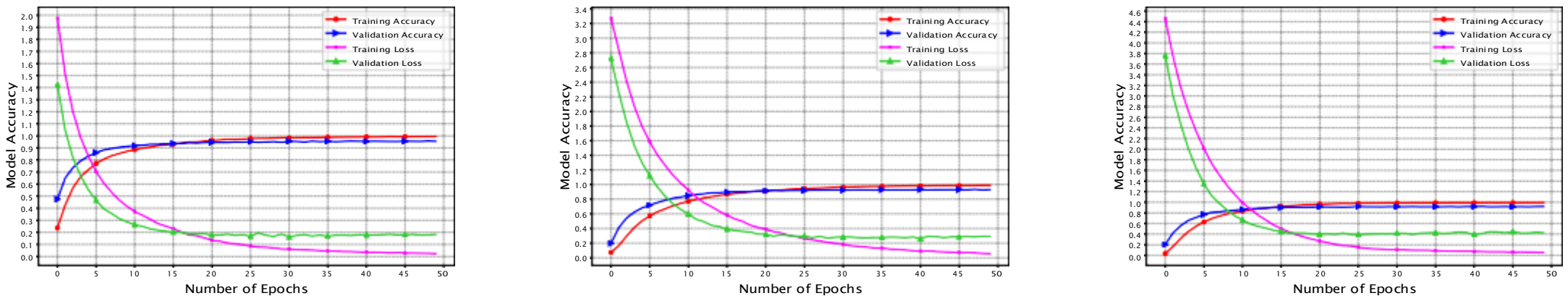

(a) (b) (c)

Figure 7: Model accuracy and loss curves for (a) 6, (b) 20, and (c) 82 classes for the three level hierarchies of Yoga-82 [5] dataset.

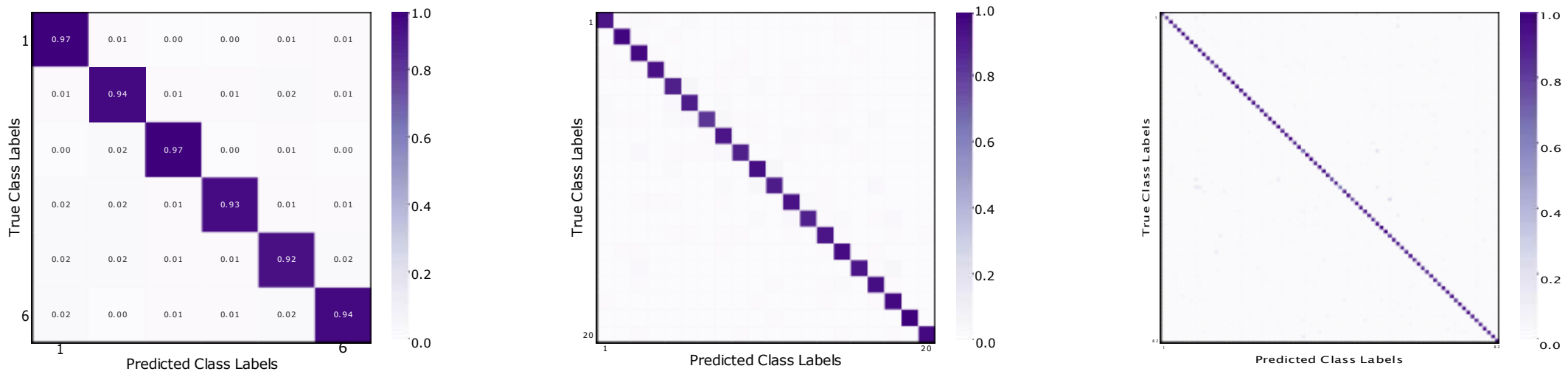

(a) (b) (c)

Figure 8: Confusion matrices for (a) 6, (b) 20, and (c) 82 classes for the three level hierarchies of Yoga-82 [5] dataset.

It is clear that the model was able to converge after 30 epochs followed by minor fluctuations. It can be noted that for the initial epochs the validation accuracy was higher than that of training. This is because of the regularisation used while training. 40% of the features are set to zero using the dropout of 0.4. However, for the testing, all of the features are used resulting in better generalization ability and robustness of the model. The best model weights were saved for evaluation on the test data.

## 4.Evaluation Parameters

This section provides a detailed analysis of the evaluation parameters. The proposed approach involves adding several dense refinement blocks to improve the pose predictions. The number of such units to be used is a crucial hyperparameter to our approach Figure 9. From the experiments, it has been observed that on each level of hierarchy, the optimum performance was after addition of 16 such blocks. It can also be noted that the addition of more units results in scaling the depth of architecture and increases the number of parameters to be trained, leading to possible overfitting on the limited data. Hence, to maintain the efficiency of yoga pose recognition and at the same time achieve optimum performance, we select 16 as the number of such units to be added.

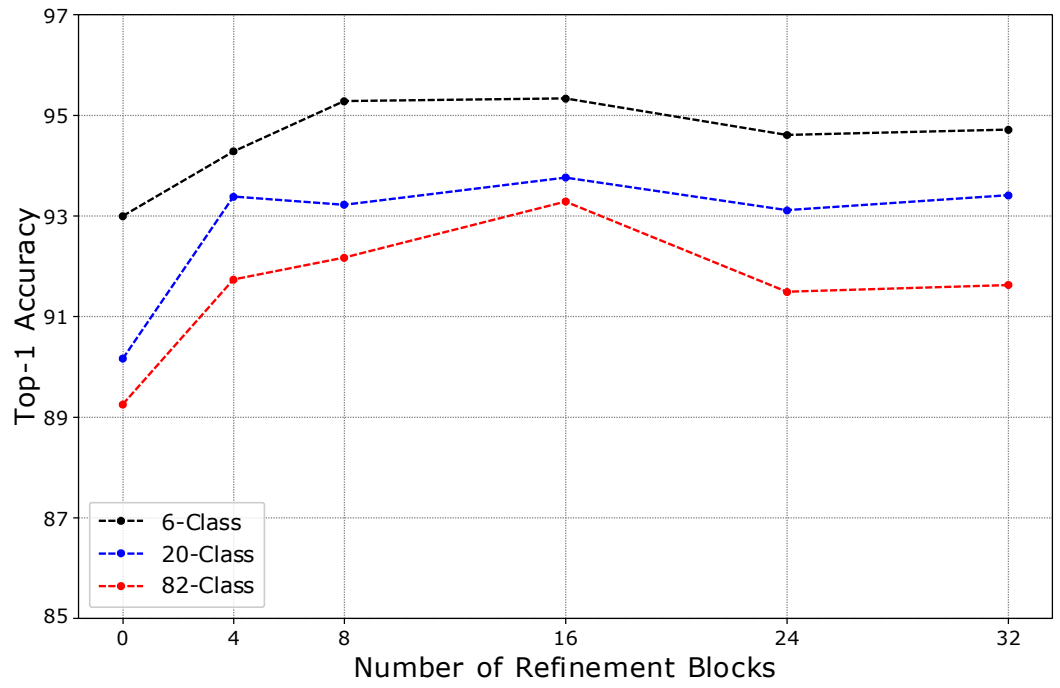

Figure 9:Performance evaluation with different number of refinement blocks.

## 5. Experimental Results

This section presents the experimental results of the proposed model on the Yoga-82 [5] dataset. It consists of four subsections. In the first subsection, dataset details are presented. In the second subsection, experimental settings are explained. In the third subsection, the performance of our proposed model is analyzed.In the fourth subsection, a detailed ablation study is presented to highlight the contribution of each module in the proposed model.Finally, in the fifth subsection, the results of our proposed model are discussed and compared with the state-of-the-art.

### *5.1. Dataset Details*

The performance of the proposed model has been evaluated on the Yoga-82 [5] dataset. This dataset is a publicly available benchmark dataset for large-scale yoga pose recognition with 82 classes. It consists of various complex yoga poses that a human body can perform. All the poses are structured into a three-level hierarchy including body positions, variations in body positions, and the actual pose names. The class labels consist of three levels with 6, 20, and 82 classes for the first, second, and third levels of the hierarchy, respectively. The dataset contains links to the images from the web for a total of 82 yoga-pose classes. The dataset has been downloaded from the links provided by the authors of the dataset. The total number of images in the dataset is 28.4k. Along with the dataset, the annotations and hierarchical class labels were provided. Many images contain *in the wild* yoga poses, along with a number of synthetic images containing silhouette poses and cartoon sketches. Also, in few of the images, multiple practitioners are present. Furthermore, images are captured from different viewing angles. Overall the dataset is challenging in terms of pose diversity and body keypoint occlusion [5]. The dataset consists of a minimum of 64 images to a maximum of 347 images of yoga poses per class.

### *5.2. Performance Evaluation*

In this subsection, the evaluation results of the proposed model are presented. As the dataset contains a three-level hierarchy of yoga poses, we present the performance of our model for the classification of all three of these levels of classes. To evaluate the proposed YPose network, we further performed various experiments on the model architecture. Extensive experiments are conducted with different CNN architectures like DenseNet [32], EfficientNet [15] for the choice of model backbone and their performances are compared to find an architecture with efficient performance. Finally, we modified the architecture of the network to obtain better representation of the input yoga pose images to get the predictions.

Table 1: Experimental results of the proposed model for the 82 classes.

| Model Variant | Precision | Recall | f1-score | Accuracy |
|---|---|---|---|---|
| YPose Lite | 0.8429 | 0.8018 | 0.8109 | 84.51% |
| YPose Network | 0.9287 | 0.9223 | 0.9241 | 93.28% |

Figure 10 shows a detailed comparison of the performance of different state-of-the-art pose estimation networks along with our proposed model on a few yoga images. In the figure, the first column presents the RGB images of the yoga pose. The columns from second to fifth demonstrates the preformance on pose estimation using OpenPose [9], HRNet [10], PifPaf [11], and Fast Pose [12]. The sixth column presents the results of ROI segmentation for the instance of yoga practitioners, which consists of a segmentation mask along with the bounding boxes using our proposed method. The last column shows the activation heatmaps learned by the proposed model on the refined images.As it can be seen from the figure, many of the state-of-the-art pose estimation methods struggle to produce accurate pose estimations. A number of keypoint detection errors were observed in the predictions. For few yoga images, not a single keypoint was detected. Also, in the case of cartoon and silhouette poses these estimation networks would catastrophically fail to predict the complex body keypoints. The activation maps of our YPose network show that the proposed model was able to learn high-level features for these complex yoga poses.

The experimental results of the proposed model are presented in the Table 1 for the 82 classes. The performance metrics are precision, recall, f1-score, and accuracy on the test split. The metrics, precision, recall, and f1-scores are calculated as macro averages. Table 3 presents the Top-1 and Top-5 accuracies for the three-level hierarchies of the test dataset.The Top-1 accuracies for six, twenty, and eighty-two classes are 95.33%, 93.38%, and 93.28%, respectively.

The six-level hierarchy classifier achieves better accuracy compared to the other two hierarchies as it comprises of simple poses of sitting, standing, balancing, *etc*. as shown in Figure 3. A small performance drop is observed for the class level hierarchy classifier, as the 82 classes of yoga are very complex in terms of intra-pose variability, inter-pose similarity and pose complexity. For the classification of 82 classes of yoga pose a training accuracy of 99.09% and validation accuracy of 92.86% was achieved. On the testing split, 93.28% accuracy was obtained. The accuracy and loss curves are plotted in Figure 7. Figure 8 illustrate the confusion matrices for 6, 20, and 82 class predictions. From the figure, it can be seen that the proposed network is able to classify each class of yoga pose correctly with very few misclassifications. The numbers on the boxes show the accuracy of that particular class.

Figure 12 presents the prediction results of our proposed model for the three-level hierarchical classes for some of the complex yoga postures. These images have diverse environments. The three-level hierarchy predictions along with their bounding boxes are displayed. As it can be infered, in some of the images the yoga practitioner is not properly visible and occupies a small portion of the image, still, the proposed model is able to correctly recognise the pose.

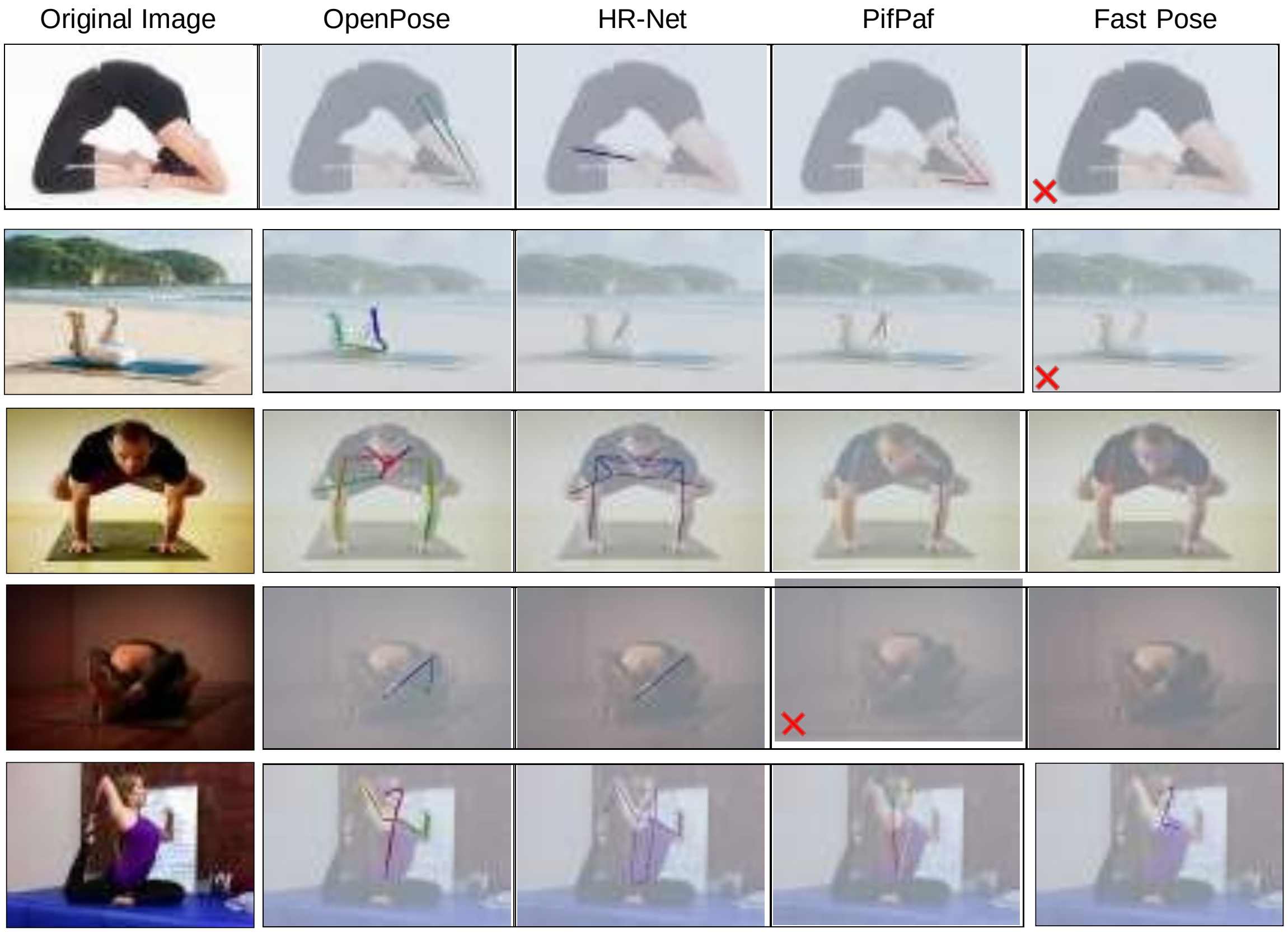

Figure 10: Yoga pose estimation using different pose estimation methods. The first, second, third, and fourth columns presents yoga poses estimated using OpenPose [9], HRNet [10], PifPaf [11], and Fast Pose [12], respectively. It is seen from the figure that most pose estimation based frameworks fail to correctly predict the keypoints, especially in the case of self-occlusion. ✘ shows not a single keypoint detected.

Table 2:Performance of the EfficientNet baselines on Yoga-82 [5] dataset. Accuracies for variant B0, B4 and B5 of EfficientNets [15] are presented. We pass refined yoga images to each baseline and reported prediction results are on 82-class hierarchy of Yoga-82.

| Baseline | Top-1 | Number of Parameters |
|---|---|---|
| EfficientNet B0 | 86.76% | 4.11 M |
| EfficientNet B4 | 89.25% | 17.69 M |
| EfficientNet B5 | 89.46% | 28.51 M |

### *5.3.Ablation study*

This section presents a detailed study of contribution from each module used in the proposed approach. To demonstrate the effectiveness of refinement blocks we evaluate the performance after removing them from the network. From the Table 4 it is observed that removing the refinement blocks leads to a significant performance drop and the CNN backbone struggles to learn representations of complex yoga poses. The refinement blocks are able to learn fine-grained representations of self-occluded and other complex yoga poses. We also conduct experiments to show the contribution of ROI extraction module. From the results it is clear that using person instance segmentation to first extract the region of interest improves the results to some extent. This is largely due to the challenging yoga scenerios where the practitioner is located in only a small region of image. The information of the practitioner's environment is not relevant to the recognition task. Finally, we achieve best performance after adding both ROI and refinement blocks to the model.

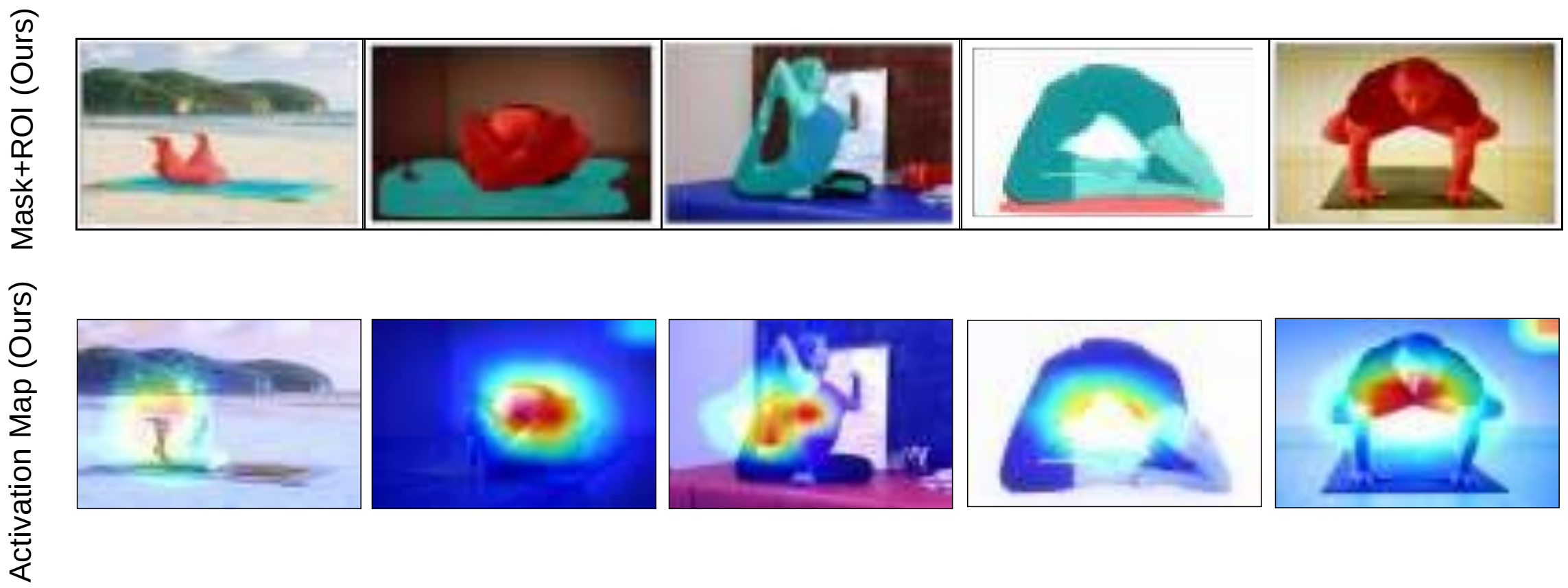

Figure 11: The first row presents the ROI along with segmentation masks using the proposed approach. The last column presents activation maps on the refined images. Our approach aims to directly recognise the poses by learning the difficult representations of yoga pose.

Table 3: Top-1 and Top-5 accuracies of the proposed model for 3 level hierarchy of the Yoga-82 [5] dataset.

| Level of Hierarchy | Top-1 accuracy | Top-5 accuracy |
|---|---|---|
| Six Class Hierarchy | 95.33% | 99.89% |
| Twenty Class Hierarchy | 93.38% | 98.37% |
| Eighty Two Class Hierarchy | 93.28% | 98.04% |

To further study the contribution of refinement blocks and the generalization ability of YPose network, we change the CNN backbone to MobileNet-V2, which previously achieved a baseline of 71.11% and 88.50% and this as Top-1 and Top-5 accuracies, respectively (Table 5). In the next step, we add 16 refinement units to obtain accurate predictions with Top-1 and Top-5 accuracies of 84.51% and 94.49%, respectively (Table 5). We refer this as the "Lite" variant of proposed YPose network. It is interesting to note that YPose Lite is lightweight with significantly reduced number of parameters and FLOPs, making it suitable for mobile and embedded system based yoga recognition applications.

### *5.4. Discussion and Comparison*

This section presents the discussion and comparison of the performance of our approach with the state-of-the-art results. The experiments were conducted involving multiple network architec-tures and evaluated the proposed approach by training each of the models.Table 5 shows a detailed comparison between the performance, the number of parameters and FLOPs of the proposed model with the state-of-the-art.For the choice of CNN backbone we studied the architecture of Efficient-Net [15]. With a lesser number of parameters, EfficientNet B0 achieves good performance. As we move from variant B0 to B7 of EfficientNet backbones the number of trainable parameters, model size, width, resolution, and depth scaling increases, consequently resulting in performance improvements.However, it was observed that for the deeper variants of EfficientNet, the accuracy does not improve by significant values Table 2. From B0 to B5, as subsequent improvement can be observed. Keeping in mind the number of parameters of the state-of-the art on the work of Verma *et al*. [5] and to lower the computational cost of network, we selected variant B4 of EfficientNets as the backbone. The proposed YPose network which is based on EfficientNet B4 followed by 16 refinement blocks was trained for predicting the 82 poses of yoga. The model has approximately 22.68 million parameters which are comparable with the models implemented in the earlier works. The network achieves a state-of-the-art accuracy of 93.28% for prediction of 82 classes of yoga pose.

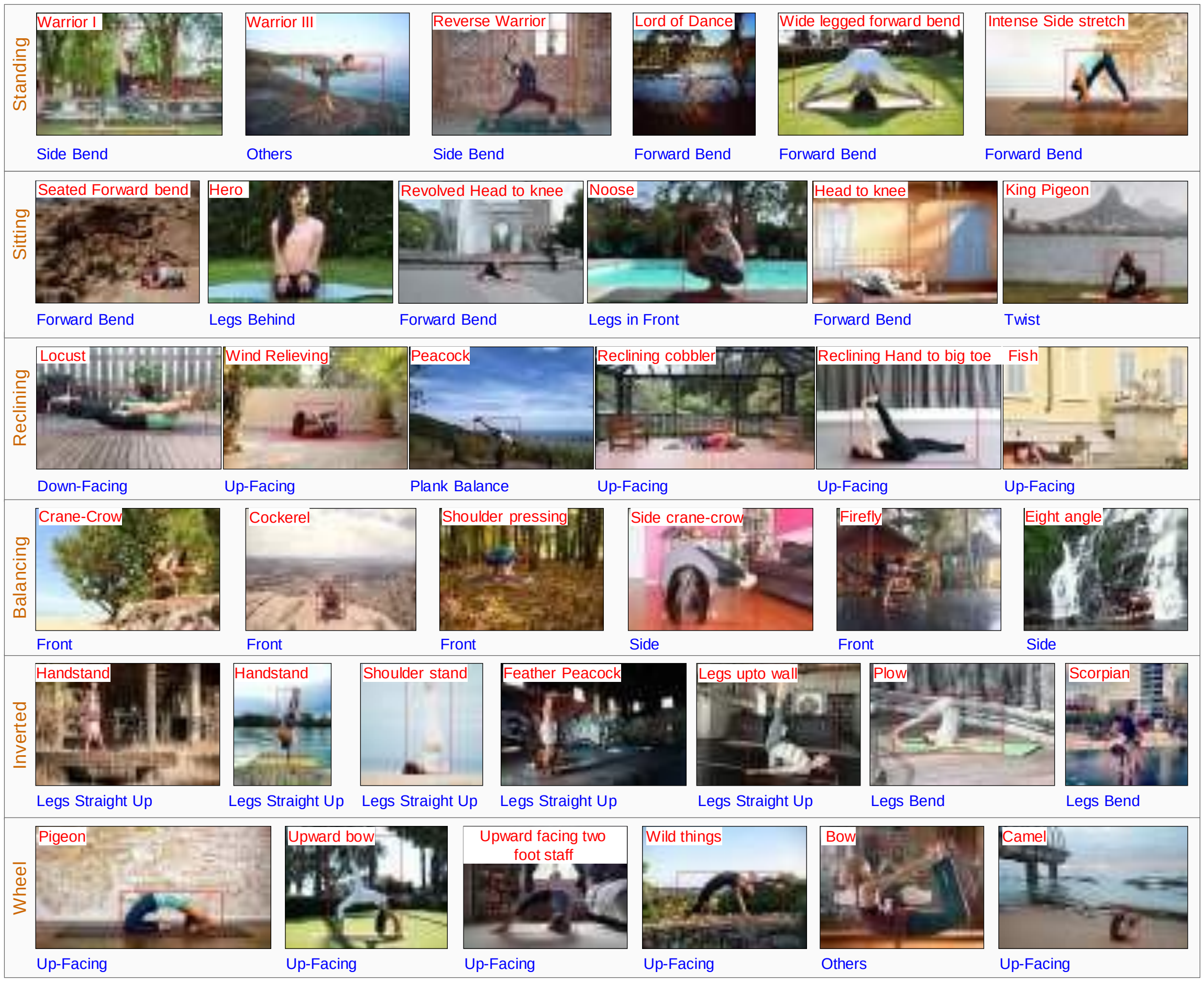

Figure 12: Three level hierarchal class prediction results for some of the complex yoga pose images performed in diverse backgrounds. The bounding frames represent the ROIs detected. The six level, twenty level, and 82 class level hierarchy is shown inorange,blue, andredcolors, respectively.

Table 4:Ablation study highlighting the contribution of different module used in the proposed model.

| Method | Top-1 accuracy (6-class) | Top-1 accuracy (20-class) | Top-1 accuracy (82-class) |
|---|---|---|---|
| ROI+Backbone | 92.99 % | 90.16 % | 89.25 % |
| Backbone+Refinement Blocks | 93.85% | 92.32% | 89.86 % |
| ROI+Backbone+Refinement Blocks | 95.33% | 93.76 % | 93.28 % |

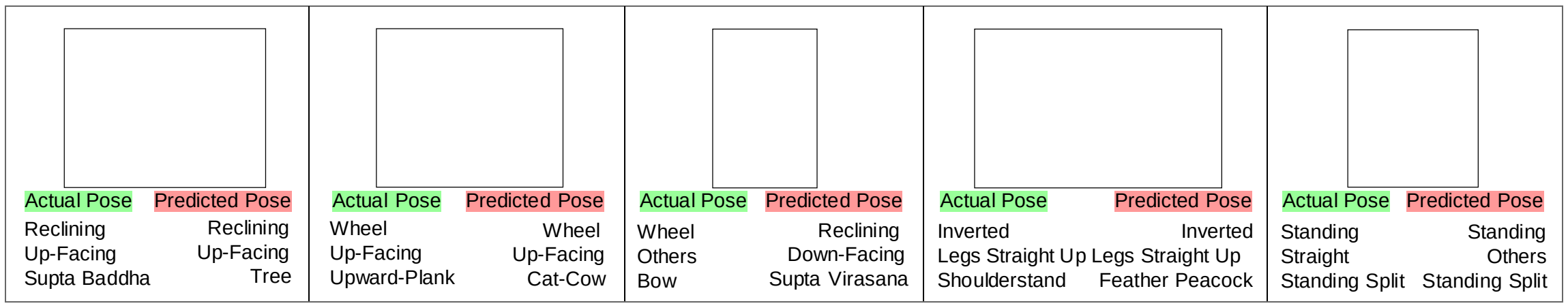

Figure 13:Some of the cases where the proposed network fails to correctly recognise the yoga pose. The ground truth and predicted hierarchical yoga poses are marked with green, red, respectively. However, the predictions on most of the pose hierarchies are accurate. The predictions on last two columns show that the approach is robust to silhouette and cartoon poses as well.

The proposed model has achieved Top-1 and Top-5 accuracies of 93.28% and 98.04%, respectively. While the best results of the work of Verma *et al*. [5] were 79.35% and 93.47% as Top-1 and Top-5 accuracies, respectively. The proposed model outperforms the earlier works with a large margin of approximately 13.9% at the same time the number of parameters and FLOPs are also comparable. From the results, it is observed that our model achieves better performance.

It is interesting to observe the cases where the proposed network fails to correctly recognise the yoga pose. Figure 13 presents some of the failure cases on the hierarchical poses of the Yoga-82. It can be observed that the proposed network is able to correctly recognise most hierarchies, even for silhouette and cartoon poses.

## 6. Conclusion

In this paper, a novel approach has been proposed for recognizing yoga poses from a single RGB image. The proposed model consists of four main steps. In the first step, ROI segmentation was performed, to detect the instance of the person in the image. This was followed by ROI Align to generate the corresponding segmentation masks and bounding boxes. In the second step, the refined images obtained from extracting the ROI extracted were passed to a CNN backbone of EfficientNet [15] to perfrom feature extraction.In the third step, several dense refinement blocks were added for extracting more diverse features, specific to complex yoga postures. Finally, in the fourth step, global average pooling and fully connected layers were applied to obtain the class predictions. The proposed network had been tested on the Yoga-82 dataset [5] and outperforms the previous state-of-the-art with a significant margin of approximately 13.9%.This work serves as a demonstration of yoga pose recognition in single images where temporal context. A similar approach can also be used for action recognition in single RGB images. In future works, data augmentation techniques can also be applied to further improve the results. Also, better segmentation approaches can be proposed for single image-based yoga pose recognition or action classification.

Table 5:Comparison with the state-of-the-art results on the Yoga-82 [5] dataset. **Blue** represents the previous state-of-the-art. **Red** denotes the best results.

| Reference | Method | Parameters | GFLOPs | Top-1 Accuracy | Top-5 Accuracy |
|---|---|---|---|---|---|
| Baseline | DenseNet201 | 18.25 M | 4.29 | 74.91% | 91.30% |
| | DenseNet169 | 12.6 M | 3.36 | 74.73% | 91.44% |
| | DenseNet121 | 7.03 M | 2.83 | 73.48% | 90.71% |
| | ResNet50 | 23.70 M | 3.86 | 63.44% | 82.55% |
| | ResNet101 | 42.72 M | 7.57 | 65.84% | 84.21% |
| | ResNet50-V2 | 23.68 M | 3.97 | 62.56% | 82.28% |
| | ResNet101-V2 | 42.69 M | 8.28 | 61.81% | 82.39% |
| | MobileNet | 3.29 M | 0.55 | 67.55% | 86.81% |
| | MobileNet-V2 | 2.33 M | 0.27 | 71.11% | 88.50% |
| | ResNext50 | 23.15 M | 4.29 | 68.45% | 86.42% |
| | ResNext101 | 42.26 M | 8.10 | 65.24% | 84.76% |
| Verma *et al.* [5] | Variant 1 | 18.27 M | 4.29 | **79.35%** | **93.47%** |
| | Variant 2 | 18.27 M | 4.29 | 79.08% | 92.84% |
| | Variant 3 | 22.59 M | 5.12 | 78.88% | 92.66% |
| Ours | YPose Lite | **6.30 M** | **0.50** | 84.51% | 94.49% |
| | **YPose Network** | **22.68 M** | **4.43** | **93.28%** | **98.04%** |

## Acknowledgements

Authors would like to thank anonymous reviewers and our parent organizations for extending their support for the betterment of the manuscript. We appreciate the assistance provided by CSIR, India. Also, we would like to acknowledge Manisha Verma *et al*. for making their dataset (Yoga-82) publicly available.

## Conflict of Interest

The authors declare that they have no conflict of interest.